\documentclass{bmvc2k}

\def\thefootnote{*}\footnotetext{These authors contributed equally to this work.}

\usepackage{multirow}
\usepackage{multicol}
\usepackage{array}
\usepackage{makecell}
\usepackage{hhline}
\usepackage{ctable}
\usepackage{float}

\usepackage{tabularx,booktabs}
\newcolumntype{Y}{>{\centering\arraybackslash}X}

\title{Personalized One-Shot Lipreading for an ALS Patient}

\addauthor{Bipasha Sen\thefootnote{}}{bipasha.sen@research.iiit.ac.in}{1}
\addauthor{Aditya Agarwal\thefootnote{}}{aditya.ag@research.iiit.ac.in}{1}
\addauthor{Rudrabha Mukhopadhyay}{radrabha.m@research.iiit.ac.in}{1}
\addauthor{Vinay Namboodiri}{vpn22@bath.ac.uk}{2}
\addauthor{C V Jawahar}{jawahar@iiit.ac.in}{1}

\addinstitution{
 CVIT, IIIT Hyderabad,\\
 Telangana, India
}
\addinstitution{
 University of Bath, \\
 Claverton Down, \\
 Bath, England
}

\runninghead{SEN ET AL.}{Personalized One-Shot Lipreading for an ALS Patient}

\begin{document}

\maketitle

\begin{abstract}

Lipreading or visually recognizing speech from the mouth movements of a speaker is a challenging and mentally taxing task. Unfortunately, multiple medical conditions force people to depend on this skill in their day-to-day lives for essential communication. Patients suffering from `Amyotrophic Lateral Sclerosis' (ALS) often lose muscle control, consequently their ability to generate speech and communicate via lip movements. Existing large datasets do not focus on medical patients or curate personalized vocabulary relevant to an individual.  Collecting large-scale dataset of a patient, needed to train modern data-hungry deep learning models is however, extremely challenging.
% However, datasets and methods neither focus on such personalized lipreading nor use datasets collected from medical patients adhering to their needs. 
In this work, we propose a personalized network to lipread an ALS patient using only one-shot examples. We depend on synthetically generated lip movements to augment the one-shot scenario. A Variational Encoder based domain adaptation technique is used to bridge the real-synthetic domain gap. Our approach significantly improves and achieves high top-5 accuracy with $83.2\%$ accuracy compared to $62.6\%$ achieved by comparable methods for the patient. Apart from evaluating our approach on the ALS patient, we also  extend it to people with hearing impairment relying extensively on lip movements to communicate. %Lastly, we evaluate our technique on a person without disability. Our model is currently deployed privately for use by the ALS patient.%\footnote{The ALS patient and his family have consented to the use of his pictures in this work.}. 

\end{abstract}

%-------------------------------------------------------------------------
\section{Introduction}

\label{sec:intro}
Lipreading is the skill of recognizing speech visually from a person's lip movements. Humans naturally rely on lipreading to discern speech, especially in crowded and noisy environments~\cite{Rosenblum2007}. 
%Babies are also known~\cite{infants} to intently observe the lip movements of people as they start learning to speak. 
It is the fundamental mode of communication for many people, such as (1) those suffering from medical conditions such as Amyotrophic Lateral Sclerosis (ALS) - leading them to lose their voice~\cite{Zarei2015, Masrori2020}, or (2) those with hearing impairment - making it difficult for them to produce proper voice. In such cases, talking to a person without voice may need you to lipread them to understand the spoken words.
%read their mouth movements to understand the spoken words.
%On the other hand, a hearing impaired person may observe the speaker's mouth movements to either disambiguate or recognize the spoken words.

\begin{figure}[t]
\begin{center}
\includegraphics[width=\linewidth]{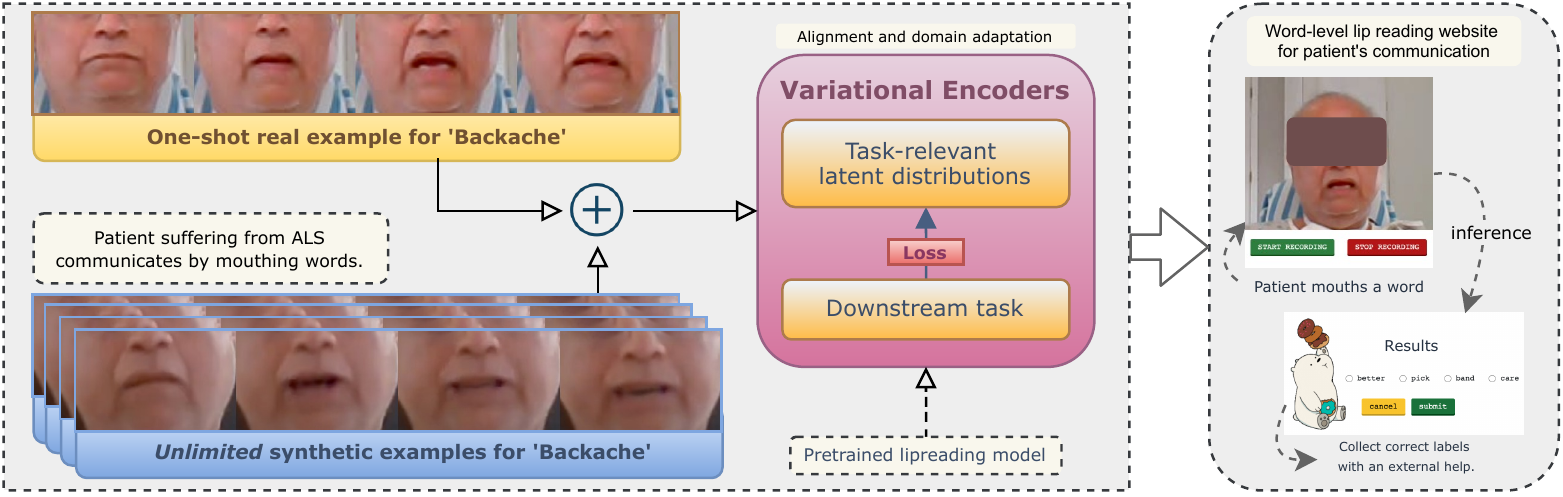}
\end{center}
\caption{We propose a personalized one-shot lipreading framework to tackle a real-world medical challenge of lipreading a patient suffering from ALS$^1$. The patient communicates primarily by mouthing the words. In our case of limited available real data, we use a combination of synthetic data augmentation technique and a domain adaptation technique called Variational Encoders to build a robust word-level lipreading model for the patient.}% We deploy the best model privately for the patient's use and extend our approach to four other speakers. }
%   \caption{We propose a framework for personalized one-shot lipreading to tackle a real-world medical challenge of lipreading a person suffering from ALS relying on mouthing words as his primary mode of communication.  We use synthetic data to augment one-shot examples collected from him and Variational Encoders to bridge synthetic-real domain gap. This results in improved performance and a high top-5 accuracy.
% %   Using synthetic data and Variational Encoders for domain adaptation helps us achieve a high top-5 accuracy. 
% We deploy the best model privately for the patient's use and extend our technique to four other speakers. }
   
%   We face the challenge of extremely low real training data and use synthetic data to augment the real data. A domain adaptation network - Variational Encoders is used to train a robust classifier using both types of data. This model is deployed through a website that lipreads the patient and displays the inferred works back to him.}
   
%   We train an architecture for lipreading a patient suffering from `Amyotrophic Lateral Sclerosis' with extremely limited number of real training examples. To deal with the data scarcity, we generate synthetic data and train a classifier on it. We specifically use a Variational Encoder module to adapt the real and synthetic domains. We also create and deploy a webapp }
\label{fig:teaser}
\end{figure}

\def\thefootnote{$^1$}\footnotetext{The ALS patient and his family have consented to the use of his pictures in this work.}

Lipreading is mentally taxing and can affect the communication quality. For instance, speakers with hearing impairment may lack holistic audio feedback~\cite{deafspeechdisability} and ALS patients may have lesser control over their mouth muscles~\cite{Beukelman2011, Makkonen2017}. This may cause them to have irregular and unreliable mouth movements making it difficult for people to lipread them. Applications and automated algorithms capable of lipreading a person can thus significantly improve the day-to-day communication of people dependent on lipreading. 
% Motivated by this need, we takle the real-world medical challenge of lipreading people 
% \footnotetext{\url{http://escolasimaohess.blogspot.com.br/}}
% Lipreading is a mentally taxing task that can affect communication quality. Applications and automated algorithms capable of lipreading a person will significantly improve the day-to-day communication of people dependent on lipreading. 
% \subsection{Challenges of Lipreading}
% Even though an essential skill, lipreading is fundamentally ambiguous due to the presence of homophenes - multiple words corresponding to identical lip movements. For example, words like \underline{ma}n, \underline{pa}n, and \underline{ba}n, although sound different, have identical lip movements. Such challenges make the task of lipreading mentally taxing and affect communication quality~\cite{DANN}. Applications and automated algorithms capable of lipreading a person will significantly improve the day-to-day communication of people dependent on lipreading.  
Motivated by this need, we tackle the real-world challenge of lipreading a patient suffering from ALS and people with hearing impairment. ALS is a progressive nervous system disease that affects nerve cells in the brain and spinal cord, causing loss of muscle control~\cite{Zarei2015}. An ALS patient may lose their voice and rely solely on mouth movements for communication~\cite{hopwood_2021, nationaldeafchildrensociety}. 
% Since, lipreading in itself is a complicated task, we focus on a comparatively simpler task - word-level lipreading in this work.
% In this case, the people around the patient need to resort to lipreading to communicate with them.
% %They may also have to rely on making lip movements to communicate with another person on a daily basis.  
% Since, lipreading in itself is a complicated task, we focus on a comparatively simpler task - word-level lipreading in this work. Our approach involves classifying lip movements into classes of different words with a given vocabulary. We also show the effectiveness of our approach by lipreading people with hearing impairment.

% \subsection{Current Works and Datasets}
% \subsection{Current Works and Limitations}

%Today's deep learning techniques are data hungry. 
\textbf{Current Works and Limitations: }Current deep learning techniques are inherently data-hungry. Collecting large amounts of data specifically from a patient is, however, not an option. Mouthing words is a tiring maneuver for people suffering from ALS, and thus a patient undergoes physical and mental stress during such data collection exercises. Manually labeling words mouthed by a person is time-consuming. 
%requiring expensive manual labour. 
It is thus crucial to use the minimum amount of manually labeled data to build lipreading models that can work well on a person. 
%given patient. 

% In this work, we formally tackle the challenge of few-shot lipreading. personalized one-shot lipreading for an ALS patient and also explore the same for four other speakers. 

Recent years have seen much progress in word-level~\cite{martinez2020lipreading, feng2020learn, Torfi_2017} and sentence-level~\cite{Afouras_2019, xu2018lcanet} lipreading. Oxford's Visual Geometry Group released large-scale in-the-wild datasets such as Lipreading in the Wild (LRW)~\cite{Chung16} and Lipreading sentences (LRS)~\cite{Afouras18c, Afouras18d} consisting of $1000+$ speakers. 
%These datasets comprise a large in-the-wild corpus of spoken words by more than 1000 speakers. 
% They capture large natural variations in terms of speaking style, pose, and lip definition of the speakers.
LRW, the most relevant dataset to our task, is a word-level lipreading dataset made of $1000$ examples for $500$ English words but turns out to be somewhat limiting: (1) The speakers in the dataset do not have any speaking disability thus making perfect mouth movements. (2) It is curated by cropping words from long speech segments resulting in fast-paced speech with co-articulation in the videos. (3) It contains a large amount of head motion and variations like the different characteristics of the mouth region, both of which are unnecessary for lipreading specific medical patients. 
%Lastly, speakers with hearing impairment may lack holistic audio feedback~\cite{deafspeechdisability} and ALS patients may have lesser control over their mouth muscles~\cite{Beukelman2011, Makkonen2017}, causing them to speak slowly and make unreliable mouth movements. 
Thus, SOTA models like LipReading without Pains (LRwP)~\cite{feng2020learn} and Lipreading using Temporal Convolutional Networks (LTCN)~\cite{martinez2020lipreading} trained on LRW do not directly adapt to speakers with speaking disabilities.

\begin{figure}[t]
\begin{center}
\includegraphics[width=\linewidth]{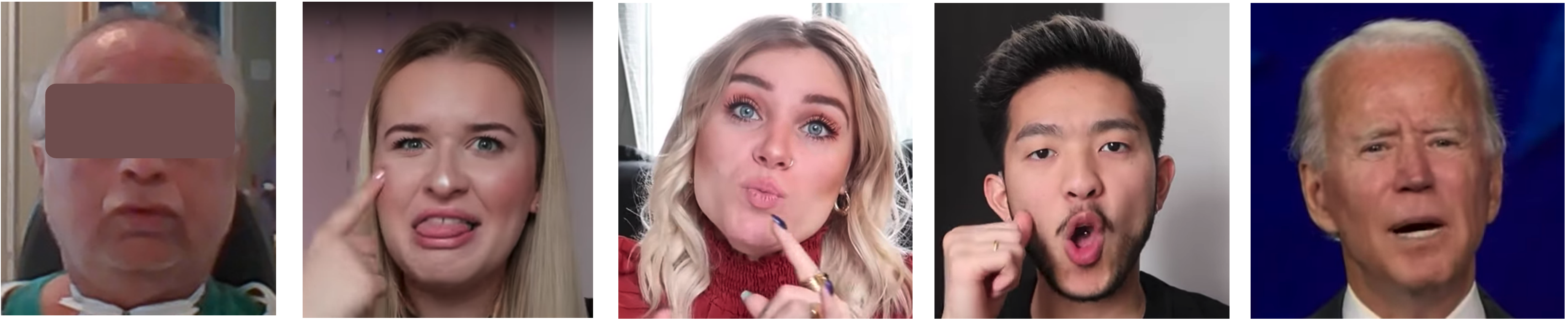}
\end{center}
  \caption{Speakers in our study. In left to right order, the first speaker suffers from ALS and relies solely on lip movements. The next two speakers primarily use sign language while making imperfect lip movements. Next speaker uses deaf speech along with sign language for daily communication. The following speaker is the 46th president of the USA, Joe Biden.}

\label{fig:speakers}
\end{figure}

LRW only supports a limited pre-curated vocabulary missing out on medically essential words like `nauseous' or `backache'. 
It also lacks a personalized vocabulary relevant to a person's daily communication. 
Deploying systems to enable a persons' communication would need highly accurate models on their specific lip movements for their particular vocabulary.
%This would cause change in domain distribution and label space between people.} 
The problem of personalized lip reading has also been explored in ~\cite{prajwal2020learning}. According to~\cite{prajwal2020learning}, lip movements vary across speakers. Observing a single speaker for an extended period could lead to better speaker-specific lipreading models. They collect $\sim$20 hours of data per speaker to train a personalized lipreading model generating speech purely from an individual speaker's lip movements. Collecting such a large dataset is, however, not always an option.

Lack of medical data has been studied~\cite{tajbakhsh2019surrogate, Frid_Adar_2018, gyawali2020semisupervised} widely in the past. 
Taking inspiration from these, we formally tackle one-shot lipreading in a personalized setting.
We first synthetically generate data using a SOTA lipsync network~\cite{wav2lip}. We then use a SOTA lipreading network~\cite{feng2020learn} as our backbone and use synthetically generated data along with very limited real examples to train a word-level lipreading network. 
%network to classify lip movements in word classes. 
Our approach includes an important domain adaptation step using a novel network -- Variational Encoders -- a modified VAE for bringing a vast number of synthetic examples closer to the real domain using only one real instance per class. We specifically tackle the use case of an ALS patient and also explore the same for four other speakers (refer to Fig.~\ref{fig:speakers}) including the 46th President of the United States, Joe Biden, as an additional example to show that our approach can easily be extended to speakers with no disability.
%We train separate models for each speaker (refer to Fig.~\ref{fig:speakers}) that include the 46th President of the United States, Joe Biden, as an additional example to show that our approach can easily be extended to speakers with no disability. 
%Finally, we evaluate our model on a publicly available dataset LRS3 - modified to suit our setting. 
Our contributions in this work are threefold: 

\begin{enumerate}
    \item We tackle a real-world medical challenge of lipreading speakers with ALS and hearing impairment by developing highly accurate personalized models for each speaker.
    \item We propose Variational Encoders, a novel network-based on VAE. Instead of autoencoding, they exploit the loss of the downstream task for generating task-relevant latent distributions. The learned distributions are then used for domain adaptation.
    \item To the best of our knowledge, we are the first to propose lipreading in one-shot setting. In this vein, we curate a medical dataset involving speakers with medical conditions.
    %We further tackle lipreading in a personalized label space.} 
    % To the best of our knowledge, we are the first to propose a method for lip reading medical patients with only a single example available per word during training. We also concentrate on developing personalized models aimed at achieving high accuracy for particular speakers....
    % required to bring the real and synthetic domains closer.}
    %thus eliminating the need for input reconstruction as in traditional VAE 
    % introduces target domain variations by sampling from the learnt latent space. }
    % \item We solve a real-world challenge of an ALS patient's ability to communication by building a word-level lipreading model with only one-shot real-world examples. % using a combination of synthetic data, pseudo labels and \textit{variational encoders}.
    % \item We discuss the framework for personalized lipreading on completely new classes without the need of large-scale labeled real data by exploiting SOTA lipreading models.
    %and also introduce a modification of an existing dataset - LRS3 - suitable for lipreading in one-shot setting. 
\end{enumerate}

\section{One-shot personalized lipreading framework}

\begin{figure*}
\begin{center}
\includegraphics[width=\linewidth]{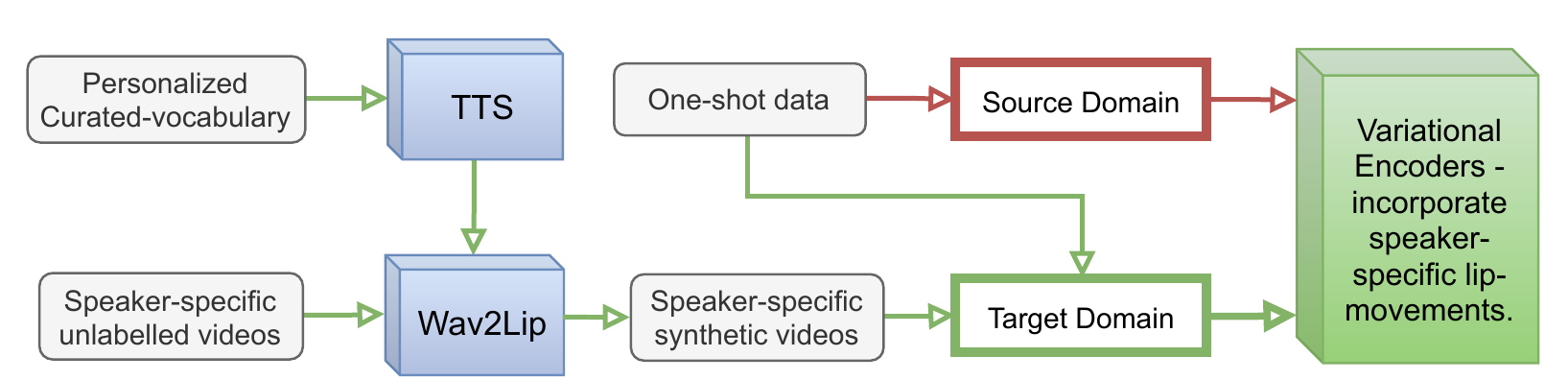}
\end{center}
   \caption{Personalized lipreading - We start by curating personalized vocabulary and collect unlabeled videos for each speaker. TTS models are used to generate speech utterances for the curated words. The speaker's unlabeled videos and the generated speech utterances are given as input to Wav2Lip that generates synthetic data to augment one-shot data. Variational Encoders then use the synthetic and the one-shot data to train the model for lipreading.}
   
%   A combination of speaker-specific synthetically generated videos and a single labeled video for each of the curated words is used. The synthetic data is generated using Wav2Lip - a face generation model capable of generating realistic mouth movements on a face video given any speech utterance. Unlabeled videos are collected for each speakers and TTS models are used as a pseudo for the speaker's voice for generating the synthetic video. In the Variational Encoder, real and synthetic data is combined to act as the source domain while the single real examples act as the target domain. }
\label{fig:pipeline}
\end{figure*}

As mentioned previously we aim to build a personalized lip-reading model for each speaker using only single real examples. %In our work, we use only one example per class for training the model.  
Fig.~\ref{fig:pipeline} presents the pipeline for personalized lipreading.

The use of synthetic data to augment low data in the medical domain~\cite{tajbakhsh2019surrogate, Frid_Adar_2018, gyawali2020semisupervised} has gained traction with improving generative models. Similarly, we augment the one-shot examples collected for each speaker by generating synthetic data for each of them. To achieve this, we use a SOTA talking face generation model, Wav2Lip~\cite{wav2lip} pretrained on the large-scale LRS2~\cite{lrs2} dataset. Given a speaker's video, Wav2Lip preserves the speaker's pose, facial expressions, and mouth characteristics like beard and skin color while modifying the speaker's lip movements according to a guiding speech. We generate word-level speech utterances using SOTA TTS models FastSpeech2~\cite{ren2021fastspeech} and GlowTTS~\cite{kim2020glowtts} as a replacement for the speaker's voice. Using these TTS models allows us to generate variations in the speech in terms of the speed of the spoken word, pitch, and energy. Additionally, we collect unlabeled face videos for each speaker which, along with the generated speech utterances, is used to create 1 hour of speaker-specific synthetic data on an average. The augmented dataset is then used to train an LRwP and LTCN based architecture for the classes curated per speaker.

% use a speaker-specific synthetically generated dataset to train an LRwP~\cite{feng2020learn} based architecture for the classes curated per speaker. To achieve this, we use a SOTA talking face generation model, Wav2Lip~\cite{wav2lip}. Wav2Lip preserves the speaker's pose, facial expressions, and mouth characteristics like beard and skin color while modifying the lip movements in a face video according to a guiding speech. We generate word-level speech utterances using SOTA TTS models FastSpeech2~\cite{ren2021fastspeech} and GlowTTS~\cite{kim2020glowtts} as a replacement for the speaker's voice. Using TTS models allows us to generate variations in the speech in terms of the speed of the spoken word, pitch, and energy. 
% %To generate the synthetic data, we collect unlabeled videos. These are the videos recorded of the patient post-ALS without any manual supervision. For the other speakers, the videos are randomly selected from their respective YouTube channels. 
 %Speaker-specific synthetic data helps the model learn visual components such as the lip definition and the complexion of the speaker.

\label{section:synth_data_framework}

A combination of the speaker-specific synthetic videos and a single real video per class are used to train our model.
% We exploit LRwP pretrained on the large-scale dataset LRW for converge faster on our speakers by using the pretrained weights as the initial weights for our models. 
% LRW being a large-scale dataset capturing wide variations in speaker characteristics, we use LRwP to exploit the powerful lip encoding for faster convergence on our speakers. We thus use the pretrained weights of LRwP, having the highest accuracy on LRW, as the weight initialization for all of the models in our experiments. 
The synthetic data helps the model learn the general underlying word-level characteristics for the new classes. However, the properties of personal style of lip-movements for a word -- could be because of the medical condition -- is not captured in the synthetic dataset. 
%This is because all the models used for generating the synthetic data -- FastSpeech2 and GlowTTS trained on LibriTTS, Wav2Lip trained on LRS2 -- are trained on datasets of speakers with no speaking disability. 
We utilize the one-shot examples for introducing the properties of personal speaking style in the model. Single examples per class are however, not enough to capture the underlying style variations of a speaker. A person may not utter the same word, exactly, each time. To tackle this, we use our novel approach -- Variational Encoders.%, as explained below.

% the purpose of finetuning LRwP. Variational Encoders hallucinate variations of the captured real examples by generating speaker specific word-level latent distributions. As a result the source encoders learns to generate domain invariant features against pseudo variations of the real input examples. At the same time, the classifier is trained on variations on synthetic and real data thus making it more robust. For faster convergence on our speakers, we use the pretrained weights of LRwP as the initial weights for our model. 

% \section{VEDA: Variational Encoders for Domain Adaptation}
\subsection{Variational Encoders: Mapping words into distributions}
\label{section:veda}

Deep learning suffers from the fundamental challenge of source-target domain shift - a model trained on a given dataset (source domain) performs poorly on the test examples (target domain). The target domain may lack the amount of labeled data needed for training or fine-tuning a model. 
Recent techniques include adversarial networks to generate domain invariant features using adversarial losses~\cite{DANN, FADA, ADDA} and
Variational Auto Encoders (VAE)~\cite{hsu2017unsupervised, takahashi2020partiallyshared} to generate a joint latent distribution across domains with KL divergence and reconstruction loss. Unlike adversarial loss~\cite{DANN} that may quickly become lopsided~\cite{DAOSL}, VAEs use a distance-based metric to incorporate domain invariance. The reconstruction loss in VAE ensures that latent representations preserve important domain characteristics. However, using a decoder to reconstruct the input accurately is a non-trivial task, especially for videos needing spatial and continuous temporal reconstruction. We propose Variational Encoders, a modified VAE that uses the loss of the downstream task to generate task-relevant latent distributions. 

\begin{figure}[t]
\begin{center}
\includegraphics[width=0.75\linewidth]{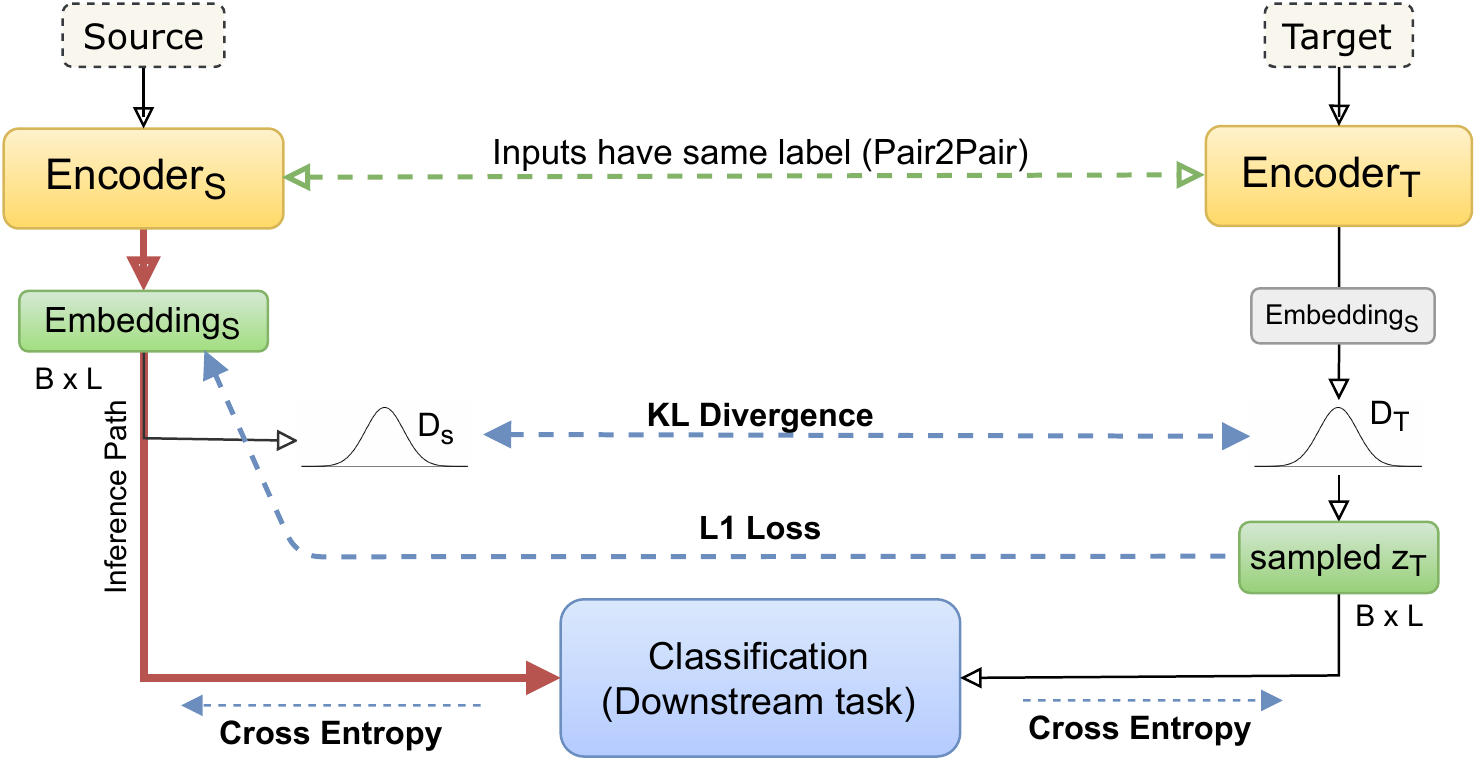}
\end{center} 
   \caption{Variational Encoders -- Encoder$_S$ and Encoder$_T$ denote the source and target encoders respectively. 
%   the source and target examples are passed through two separate encoders - Encoder$_S$ and Encoder$_T$ respectively. 
    Encoder$_T$ generates a latent distribution using the target examples. The latent distribution is trained against the downstream classification task. Encoder$_S$ aims to learn robust domain invariant features by minimizing the distance between its learned embedding and points randomly sampled from the generated target distribution. % The sampled points act as a pseudo for the missed examples in the target dataset in a one-shot setting. 
   % B is the batch size and L is the latent dimension.
    % Encoder$_S$ is used for inference. 
   %The classifier is trained against a variety of synthetic (Embedding$_S$) and real (sampled z$_T$) encodings. 
   }
\label{fig:diag_main}
\end{figure}

\textbf{How do Variational Encoders differ from a standard VAE?} Similar to VAEs, Variational Encoders (refer to Fig.~\ref{fig:diag_main}) generate latent distributions given a sample. A key difference with a standard VAE is - instead of autoencoding, it uses the final multi-class classification loss of the downstream task to generate a task-relevant latent distribution that represents the ``class" of input instead of a generic input feature. The learned distribution is used to sample a variation of the input example. We hypothesize that this sample introduces the missing speaker-specific style variation for the input class. The sampled variation encourages the encoder to see and align to the potential interpretation of the input class and learn robust representations.
%acts as a pseudo for the missing samples for the speaker-specific style variation of a given class. Using the sample, we regularize the encoder making encouraging the encoder to learn robust representations for the class. 
%We use the distribution learnt by the encoder to sample variation of the input example. The sampled variation in turn allows an encoder to see potential variation of the input example and learn more robust representations.
% We hypothesize that this makes up for the missing samples for the speaker-specific style variation of a given class in a one-shot setting. 
In summary, Variational Encoder retains all the benefits of a VAE, like generating domain invariant features while removing the complicated video reconstruction loss. 
% We explain the architecture in more detail below 

%While previous architectures train the downstream and adaptation task simultaneously, the domain adaptation is not directly influenced by the downstream task. A separate loss like KL divergence or adversarial loss is used to bring the feature space across the domains closer. 
% Even though in the previously mentioned architectures, the downstream task is trained simultaneously with the adaptation task, the domain invariant features are not directly influenced by the downstream task but is encouraged through a separate loss such as the KL divergence or adversarial loss.
%, which may cause the network to learn a sub-optimal latent representation. 
%Unlike the previously mentioned architectures that use a separate loss such as the KL divergence or adversarial loss for domain adaptation, 
% Variational Encoders directly use the loss of the downstream task to generate optimal features for task in hand. This way, Variational Encoders retain the traditional VAE benfitis - 
% %This allows the network to retain the traditional benefits of a VAE -- 
% distance based loss to generate domain invariant features and distributions to introduce more variations. 
% At the same time, they maintain a simpler architecture eliminating the need for non-trivial input reconstruction whilst generating task-optimal features.

\subsubsection{Network Architecture}
\label{section:problem_definition_veda}

Given a source domain \textit{S} and a target domain \textit{T}, domain adaptation aims to bridge the gap between the two domains by generating domain invariant features. We denote \textit{S} and \textit{T} as,
\begin{equation}
    S = \{(x_1, y_1), (x_2, y_2), ... (x_{N_S}, y_{N_S})\}\ \textrm{and}\ T = \{(\bar{x}_1, \bar{y}_1), (\bar{x}_2, \bar{y}_2), ... (\bar{x}_{N_T}, \bar{y}_{N_T})\}
\end{equation}
where $N_S$ is the number of source samples and $N_T$ is the number of target samples. 
% where $(x_i, y_i)$ is an example from the source domain, $N_S$ is the number of source samples. 
%Broadly, domain adaptation can be divided into \textit{paired domain adaptation} where the underlying classes for the source and the target domain are overlapping and the examples in the target domain are also labeled. \textit{Unpaired domain adaptation} assumes that the target domain data is unlabeled. A third kind of technique assumes the target and source domain to have non-overlapping classes with labeled one-shot target examples~\cite{DAOSL}. In this work, we focus on \textit{paired domain adaptation} so that the source encoder can train against the class-level variations generated by the target encoder. However, we assume that the number of samples in the target domain is much lower than the source domain and rely on the \textit{Variational Encoder} to generate target domain variations. Thus, we denote \textit{T} as,

Variational Encoder is a paired domain adaptation network that assumes the target domain is labeled. However, the number of examples in the target domain is expected to be at max $k$ where $k\approx1$. Thus, we assume $N_S >> N_T$.
The adaptation network is made of two encoders - source and target encoders - 
%- source encoder for capturing the source domain and target encoder capturing the target domain; 
and a single classifier. 
%The encoders first generate an embedding of the input followed by a latent distribution. 
For a given input $\hat{x}_i$ belonging to either domain, the encoder learns to generates an embedding $e_{\hat{x}_i}$ for the input.
%denoted as $f_\theta : {\rm I\!R}^D \rightarrow {\rm I\!R}^M$, where $D$ is the input dimension, and $M$ is the dimension of the learnt embedding $e_{\hat{x}_i}$.
A distribution $p(\hat{x}_i)$ is then generated for the input denoted by $\mu_{\hat{x}_i}$ and $\sigma_{\hat{x}_i}$ from the learnt embedding $e_{\hat{x}_i}$. A random point is sampled from the generated distribution denoted as $z_{\hat{x}_i} = \mu_{\hat{x}_i} + \sigma_{\hat{x}_i} \bigodot \epsilon$.

The encoder network for both domains are identical. Each encoder generates the embedding $e$, and the latent distribution $p$. To introduce domain invariance, we use KL divergence and L1 loss. The KL divergence is computed between the two generated distributions $p(x_i) = (\mu_{x_i}, \sigma_{x_i})$ and $p(\bar{x}_i) = (\mu_{\bar{x}_i}, \sigma_{\bar{x}_i})$. In addition, a point $z_{\bar{x}_i}$ is randomly sampled from the target distribution $p(\bar{x}_i)$. L1 loss is then applied between $z_{\bar{x}_i}$ and the source embedding $e_{x_i}$. This minimizes the distance between the source embedding against several randomly sampled points from the target distribution thus acting as a pseudo for multiple target examples.

\textbf{Gradient stopping: }The L1 Loss between the source embedding and the sampled target embedding will force the target distribution down to a single point, losing the essence of a distribution. To prevent that, we stop the gradient from flowing back through the target encoder. This way, only the source encoder is regularized against the variations sampled from the learnt target distribution, while leaving the target encoder unaffected. We denote the sampled target embedding as $z_{{\bar{x}_i}detach}$, where $_{detach}$ denotes that the sample $z_{\bar{x}_i}$ does not have any gradient. 
% Applying a distance based loss on the sampled point against the source embedding will however, force the network to bring the target distribution down to a single point. Thus, we detach from the sampled point and apply the L1 loss only against the source embedding. This forces the source encoder to regularize against multiple sampled points from the target distribution, helping it learn more robust representations for a given class, while leaving the sampled point unaffected. }
%As a result, the encoder learns to generate features invariant across the source and the target domain. 
The combined loss to bridge the two domains is given as, 
%This allows the source encoder to minimize the distance between the input source embedding against variations of the target input embedding. This acts as a pseudo for a target dataset having abundant examples. 
%The target encoder learns to hallucinate variations of the real input and the source encoder is trained against these variations such that the generated embedding by the source encoder represent domain invariant features. Thus, the combined loss to bridge the two domains is denoted as,
\begin{equation}
    \Delta_{dist} = | e_{x_i} - z_{{\bar{x}_i}detach} | - \beta \cdot D_{KL}( p(x_i) || p(\bar{x}_i)).
\end{equation}
where $\beta$ is a hyper-parameter and $i$ denotes both the samples belong to the same class.

For the downstream task of classification, we employ a common classifier that is trained on both domains at the same time. 
%A single classifier allows both the encoder to have a common base-loss helping them to generate features relevant to the common downstream task. 
A single classifier allows both encoders to have a common base-loss helping them generate features relevant to the common downstream task.
The downstream classifier also trains on large amounts of variations %for any given class 
and learns robust representations. 
%We observe that employing a single classifier for both domains helps in developing features that performs well for the single common task. 
%Employing different classifiers may lead to the classifiers giving importance to different underlying characteristics for both the encoders resulting in sub-optimal task relevant features. 
% For a downstream task such as classification, sampling from a distribution may generate unnecessary variations. Thus, during inference, we use only the embeddings generated by the encoders. The source encoder is used for the inference as it is trained for generating domain invariant embeddings. 
%We use the source encoders during the inference mode as the source encoder encounters a higher amount of variation from the source as well as the target domain.
The classifier receives a randomly sampled point $z_{\bar{x}_i}$ from the learnt distribution $p_{\bar{x}_i}$ as input for the target domain. For the source encoder, however, the classifier receives the learnt embedding $e_{x_i}$ instead of a randomly sampled latent point $z_{x_i}$. 
%This is done, as during inference, we avoid using a randomly sampled point from the distribution for classification. 
The motivation behind a target distribution is to encourage the target encoder to hallucinate variations of the single real examples available for each class. However, during inference, we want to obtain a definite point for the input to avoid any uncertainties. The classification loss is given as, 
\begin{equation}
    \Delta_{entropy} = g(e_i, y) + g(\bar{z}_i, y)
\end{equation}
where $g$ can be any classification loss such as cross entropy or negative log likelihood. $y$ denote the same label for both the source $x_i$ and target $\bar{x}_i$ input. 

The combined loss for the entire network is then given as (see Fig.~\ref{fig:diag_main}),
\begin{equation}
    \Delta_{ve} = \alpha \cdot (g(e_i, y) + g(\bar{z}_i, y)) + \gamma \cdot (| e_{x_i} - z_{{\bar{x}_i}detach} | - \beta \cdot D_{KL}( p(x_i) || p(\bar{x}_i))).
\end{equation}
where $\alpha$ and $\gamma$ are hyper-parameters. All the network components are trained end-to-end.
\section{Experiments}

% In this section, we explain the experiments done for the 5 speakers as shown in Fig.~\ref{fig:speakers}. 
%We systematically move from using the one-shot real examples to synthetically generated examples as a data augmentation strategy for training the models. We then describe the improvements in the accuracy using Variational Encoders. 

% We systematically move from using one-shot real resources to using synthetically generated data as a data augmentation strategy for model training. We then showcase the ability of Variational Encoders to bring the synthetic examples and one-shot real resources for training a robust classifier. For faster convergence, we use the pretrained weights of LRwP as the initial weights for all models in our experiments. This enables the model to exploit the powerful lip encoding learnt by LRwP on the large-scale LRW dataset.   
%We then scrutinize the role of each component of our network through extensive ablation studies. 
% Table~\ref{tab:evaluation} presents a comprehensive view of the experiments done on each of the speakers. Speaker-1 uses a combination of deaf speech and sign language, Speaker-2 and Speaker-3 use sign language as their primary model of communication. 

%-------------------------------------------------------------------------
% \subsection{Experimental Setup}

\textbf{Dataset}: We first collect a set of unlabeled videos for each speaker that are used later to generate the speaker-specific synthetic data using Wav2Lip.
%For each of our speakers, we first collect a set of unlabeled videos that are later used for generating the synthetic data using Wav2Lip. 
Unlabeled videos of the ALS patient are recorded without any manual intervention. For other speakers, the videos are randomly selected from their respective YouTube channels. As the next step, we curate the personalized vocabulary for each of the speakers. For the ALS patient, we curate a list of $200$ words with the help of his family. We modify the existing list for the remaining speakers by removing irrelevant words and adding the most occurring keywords in the transcription of the collected videos. Splits of the curated dataset for each speaker is presented in Table~\ref{tab:datasplit}. 

\begingroup
\setlength{\tabcolsep}{3pt} % Default value: 6pt
\renewcommand{\arraystretch}{1} % Default value: 1
\begin{table}[H]
  \begin{center}
%   \begin{tabular}{ l | c | c | c | c | c | c  }
\begin{tabularx}{\textwidth}{ |l|l| *{6}{Y|} }
      \hline
      \multicolumn{2}{| c|  }{} & \textbf{Patient} & \textbf{Spk-1} & \textbf{Spk-2} & \textbf{Spk-3} & \textbf{Joe Biden} & \textbf{Total}
      \\ \hline
      \multicolumn{1}{| c  }{} & Classes & 200 & 75 & 70 & 80 & 75 & -
      \\
      \hline
      \multirow{3}{*}{\makecell{Train}} & Real & 200 & 75 & 70 & 80 & 75 & -
      \\
      & Synth & 17$\times$Real & 22$\times$Real & 22$\times$Real & 22$\times$Real & 25$\times$Real & -
      \\
      & Aug. & 2$\times$Real & 2$\times$Real & 2$\times$Real & 2$\times$Real & 2$\times$Real & -
      \\
      \hline
      Test & Real & 320 & 120 & 80 & 90 & 90 & 710
      \\ \hline
\end{tabularx}
  \end{center}
  \caption{Split up for the datasets curated for each of the 5 speakers. }
  %We conduct experiments on one-shot lipreading, hence the amount of real-training data is equal to the number of classes for each speaker. The classes across the 5 speakers may overlap.}
  \label{tab:datasplit}
\end{table}
\endgroup

We obtain two sets of manually curated real data from the ALS patient's family. We use the first set for model training and the second set as the intermediate test set. We collect additional data by deploying a website that records the patient's word-level mouthings and displays the inference on our best model (see Fig.~\ref{fig:teaser}). 
The website needs external help to start and stop recording the patient. 
The helper can then either select one of the displayed words as the correct label or manually assign the correct label. Through this exercise, we collect an additional $320$ examples. Out of these, we use 200 data points, in addition to the original 200 train data points, to train a model in a two-shot setting. We report our test results of one-shot and two-shot models on the combined set of the intermediate test set and the remaining additional $120$ examples (see Table~\ref{tab:datasplit}). To simulate the same setting, we maintain one real example for training the rest of the speakers. To generate the train and test examples, we use the transcriptions with timestamps available on YouTube for the selected videos.

\textbf{Preprocessing}: Our preprocessing steps are similar to LRwP. The lip landmarks are first detected using dlib~\cite{dlib}. The lip is then cropped out such that it is horizontally and vertically centered in the cropped image. The image is converted to gray-scale and resized to a fixed dimension of $88\times88$. 
A maximum sequence length of $64$ frames is used. 
% We follow a similar preprocessing strategy as used in LRwP. Lip landmark detection is performed using dlib \cite{?} and the lip region is cropped by adding a bounding box around the detected lip landmarks. The images are subsequently converted to gray-scale. The cropped region is resized to a fixed dimension of 96x96 and center cropping of 88x88 is applied such that the lip image is roughly centered on the cropped region. This allows the model to focus on lip movements rather than on facial features and pose variations. The input sequences are composed of variable length with a maximum length of 64 frames. 
% All the models in this experiments are trained with 
A batch size of $16$ on a multi-GPU NVIDIA GeForce RTX $2080$ setup is used. The models are trained up to $200$ epochs using a cosine scheduler and Adam optimizer, with a $3e - 5$ learning rate and a weight decay of $1e - 4$. The encoders in the experiments are adopted from LRwP that uses Resnet18 and BiGRU, and LTCN that uses Resnet18 and Temporal Convolutional Networks. 
%We also introduce horizontal flip to augment the data. 

\begingroup
\setlength{\tabcolsep}{3pt} % Default value: 6pt
\renewcommand{\arraystretch}{1} % Default value: 1
\begin{table}[t]
  \begin{center}
  \resizebox{\textwidth}{!}{%
  \begin{tabular}{| l || c | c || c | c || c | c || c | c || c | c|| c| c | c }
      \hline
       & \multicolumn{2}{ c || }{\textbf{Patient}} & \multicolumn{2}{ c ||}{\textbf{Spk-1}} & \multicolumn{2}{ c ||}{\textbf{Spk-2}} & \multicolumn{2}{ c ||}{\textbf{Spk-3}} & \multicolumn{2}{ c ||}{\textbf{Joe Biden}} & \multicolumn{2}{ c |}{\textbf{Avg.}}
      \\
      \cline{2-13}
      Experiment & top1 & top5 & top1 & top5 & top1 & top5 & top1 & top5 & top1 & top5 & top1 & top5\\
    \hline
    \multicolumn{13}{ | c | }{LRwP} \\
    \hline
    cl$-$r & 49.3 & 62.6 & 32.4 & 52.1 & 23.1 & 42.5 & 26.7 & 49.3 & 33.5 & 51.5 & 33.0 & 51.6 \\ 
    cl$-$r$+$s & 53.6 & 68.1 & 32.3 & 49.8 & \textbf{37.2} & 45.1 & 32.4 & 58.8 & 57.6 & 69.3 & 42.6 & 58.7 \\ 
    ve$-$r$+$s & 66.4 & 81.4 & \textbf{48.6} & \textbf{68.5} & 34.4 & \textbf{59.6} & 41.6 & 71.4 & 54.1 & 72.4 & 49.0 & \textbf{70.4} \\ 
    cl$-$r$+$s$+$aug & 61.3 & 75.2 & 34.6 & 51.2 & 33.4 & 40.4 & 51.2 & 63.6 & \textbf{64.5} & 75.3 & 49.4 & 61.3 \\
    ve$-$r$+$s$+$aug & \textbf{68.1} & \textbf{83.2 } & 44.3 & 62.6 & 31.2 & 50.8 & \textbf{53.4} & \textbf{73.5} & 61.4 & \textbf{77.1} & \textbf{52.2} & 69.4 \\
    % cl$-$r & 49.31 & 62.67 & 32.43 & 52.12 & 23.13 & 42.57 & 26.77 & 49.33 & 33.56 & 51.53 & 33.04 & 51.64 \\ 
    % cl$-$r$+$s & 53.62 & 68.11 & 32.34 & 49.82 & \textbf{37.25} & 45.14 & 32.46 & 58.85 & 57.64 & 69.34 & 42 & 58 \\ 
    % ve$-$r$+$s & 66.42 & 81.45 & \textbf{48.63} & \textbf{68.53} & 34.43 & \textbf{59.61} & 41.68 & 71.42 & 54.16 & 72.46 & 48 & \textbf{70} \\ 
    % cl$-$r$+$s$+$aug & 61 & 75 & 34 & 51 & 33 & 40 & 51 & 63 & \textbf{64} & 75 & 49 & 61 \\
    % ve$-$r$+$s$+$aug & \textbf{68} & \textbf{83 } & 44 & 62 & 31 & 50 & \textbf{53} & \textbf{73} & 61 & \textbf{77} & \textbf{52} & 69 \\
    \hline
    \multicolumn{13}{| c |}{LTCN} \\
    \hline 
    % cl$-$r$+$s & 65.2 & 81.1 & 39.3 & 72.0 & 14.7 & 53.3 & 17.2 & 57.8 & 24.5 & 56.6 & NA & NA
    cl$-$r$+$s & 55.2 & 67.1 & 36.5 & 52.0 & 34.8 & 44.8 & 34.4 & 59.6 & 55.4 & 67.3 & 43.3 & 58.2
    \\ 
    ve$-$r$+$s & 64.5 & 80.2 & \textbf{49.3} & \textbf{71.1} & \textbf{35.7} & \textbf{57.9} & 46.4 & 73.2 & 56.3 & 71.1 & 50.4 & \textbf{70.7}
    \\ 
    % cl$-$r$+$s$+$aug & 75.3 & 87.1 & 45.8 & 71.0 & 22.7 & 56.0 & 28.1 & 57.8 & 35.8 & 58.5 & NA & NA \\ 
    cl$-$r$+$s$+$aug & 61.9 & 74.2 & 33.2 & 49.8 & 32.4 & 39.9 & 52.5 & 66.1 & \textbf{62.1} & \textbf{76.2} & 48.4 & 61.2
    \\
    ve$-$r$+$s$+$aug & \textbf{66.8} & \textbf{81.7} & 41.7 & 60.4 & 32.4 & 48.8 & \textbf{54.9} & \textbf{75.2} & 61.9 & 75.3 & \textbf{51.5} & 68.3
    \\
    \hline
    % \midrule
    % lrwp:vae-r$+$s$+$aug & \\
    % \hline
    % ltcn:vae-r$+$s$+$aug & \\
    % \hline
    %   ALS patient & 0.49 & 0.54 & 0.62 & 0.53& 0.63 &0.68 & \textbf{0.66} &\textbf{ 0.76} & \textbf{0.81}
    %   \\
    %   Speaker-1 & 0.32& 0.43 &0.52 & 0.32& 0.41 &0.49 & \textbf{0.48} & \textbf{0.53} & \textbf{0.68} %Benny
    %   \\ 
    %   Speaker-2 & 0.23& 0.36 &0.42 & \textbf{0.37}& \textbf{0.41} &0.45 & 0.34 & 0.41 & \textbf{0.59} %Chen
    %   \\ 
    %   Speaker-3 & 0.26& 0.42 &0.49 & 0.32& 0.42 &0.58 & \textbf{0.41} & \textbf{0.60} & \textbf{0.71} %Jazzy
    %   \\ 
    %   Joe Biden & 0.33 & 0.45 &0.51 & \textbf{0.57} & 0.61 & 0.69 & 0.54 & \textbf{0.66 }& \textbf{0.72}
    %   \\ \midrule
    %   Avg. & 0.33 & 0.44 & 0.51 & 0.42 & 0.49 & 0.58 & \textbf{0.48} & \textbf{0.59} & \textbf{0.70}
    %   \\ \bottomrule
  \end{tabular}}
  \end{center}
  \caption{Evaluation of our models against each speaker reported in \%. All metrics are evaluated on the curated test set made of only real-data. Spk-1 uses a combination of deaf-speech and sign language, Spk-2 and Spk-3 use sign language for communication. Patient denote the ALS patient in our study. 
  %lrwp and ltcn denote that LRwP and LTCN is used as a backbone for the model. 
  r, s, and aug indicate real, synthetic, and augmented-real datapoints. cl indicates standard classification while VE is the proposed technique.}
  %cl indicates classification and ve indicates variational encoders as the mode of training.}}
  %Exp-1 is the baseline trained on one-shot examples. Exp-2 is trained on synthetic and one-shot examples. Exp-3 uses Variational Encoders trained on synthetic and one-shot examples.}}
  \label{tab:evaluation}
\end{table}
\endgroup

\begingroup
\setlength{\tabcolsep}{6pt} % Default value: 6pt
\renewcommand{\arraystretch}{1} % Default value: 1
\begin{table}[]
  \begin{center}
  \begin{tabular}{| l | l | c | c | c | }
    \hline
    \multicolumn{1}{ | c }{} & \multicolumn{1}{ c | }{\textbf{Experiment}} &\textbf{ top-1} & \textbf{top-3} & \textbf{top-5} \\
    \hline
    \multirow{3}{*}{\makecell{One-shot (classification)}} & only fs tts & 56.83 & 68.46 & 71.92 \\ 
    \cline{2-5}
    & only glow tts & 57.56 & 67.73 & 71.55 \\
    \cline{2-5}
    & combined & \textbf{61.38} & \textbf{73.63} & \textbf{75.26} \\
    \hline
    \multirow{2}{*}{\makecell{Two-shot}} & classification & 64.33 & 74.69 & 82.16 \\ 
    \cline{2-5}
    & variational encoders &  \textbf{71.64} & \textbf{76.54} & \textbf{89.36} \\ 
    \hline
    % \midrule
    % lrwp:vae-r$+$s$+$aug & \\
    % \hline
    % ltcn:vae-r$+$s$+$aug & \\
    % \hline
    %   ALS patient & 0.49 & 0.54 & 0.62 & 0.53& 0.63 &0.68 & \textbf{0.66} &\textbf{ 0.76} & \textbf{0.81}
    %   \\
    %   Speaker-1 & 0.32& 0.43 &0.52 & 0.32& 0.41 &0.49 & \textbf{0.48} & \textbf{0.53} & \textbf{0.68} %Benny
    %   \\ 
    %   Speaker-2 & 0.23& 0.36 &0.42 & \textbf{0.37}& \textbf{0.41} &0.45 & 0.34 & 0.41 & \textbf{0.59} %Chen
    %   \\ 
    %   Speaker-3 & 0.26& 0.42 &0.49 & 0.32& 0.42 &0.58 & \textbf{0.41} & \textbf{0.60} & \textbf{0.71} %Jazzy
    %   \\ 
    %   Joe Biden & 0.33 & 0.45 &0.51 & \textbf{0.57} & 0.61 & 0.69 & 0.54 & \textbf{0.66 }& \textbf{0.72}
    %   \\ \midrule
    %   Avg. & 0.33 & 0.44 & 0.51 & 0.42 & 0.49 & 0.58 & \textbf{0.48} & \textbf{0.59} & \textbf{0.70}
    %   \\ \bottomrule
  \end{tabular}
  \end{center}
  \caption{Accuracy of additional experiments performed on the ALS patient in \%. The test dataset used for both, One-shot and Two-shot experiments is the same. FastSpeech2 and GlowTTS are represented by fs tts and glow tts, respectively. All experiments are conducted using LRwP as the backbone on the combined real, synthetic, and augmented real datasets.}
  %Exp-1 is the baseline trained on one-shot examples. Exp-2 is trained on synthetic and one-shot examples. Exp-3 uses Variational Encoders trained on synthetic and one-shot examples.}}
  \label{tab:additional_evaluation}
\end{table}
\endgroup

\subsection{Training strategy}
Table~\ref{tab:evaluation} presents a comprehensive overview of all the experiments conducted on the 5 speakers. Spk-1 uses deaf-speech and sign language, Spk-2 and Spk-3 use sign language as their primary mode of communication while mouthing words with imperfect lip movements. Table~\ref{tab:additional_evaluation} presents additional experiments performed on the dataset of the ALS patient. 
We initialize our models with the weights of LRwP or LTCN, both of them pretrained on LRW.
% For training all the models, we use current SOTA lipreading model LRwP as the backbone and initialize it with its pretrained weights on LRW. 
We observed that using the pretrained weights leads to faster convergence. 
% use its pretrained weights on LRW for model initialization as we observed that using the pretrained weights leads to faster convergence.
% In this section, we explain the experiments done for the 5 speakers as shown in Fig.~\ref{fig:speakers}. 
%We systematically move from using the one-shot real examples to synthetically generated examples as a data augmentation strategy for training the models. We then describe the improvements in the accuracy using Variational Encoders. 

% We systematically move from using one-shot real resources to using synthetically generated data as a data augmentation strategy for model training. We then showcase the ability of Variational Encoders to bring the synthetic examples and one-shot real resources for training a robust classifier. For faster convergence, we use the pretrained weights of LRwP as the initial weights for all models in our experiments. This enables the model to exploit the powerful lip encoding learnt by LRwP on the large-scale LRW dataset.   
%We then scrutinize the role of each component of our network through extensive ablation studies. 
% Table~\ref{tab:evaluation} presents a comprehensive view of the experiments done on each of the speakers. Speaker-1 uses a combination of deaf speech and sign language, Speaker-2 and Speaker-3 use sign language as their primary model of communication. 

\textbf{Baseline (Exp-\textit{cl$-$r})}: We begin by training our model directly on the one-shot examples. Since we are the first to perform one-shot lipreading on a personalized vocabulary, we treat this model as our baseline. Table~\ref{tab:evaluation} Exp-\textit{cl$-$r} presents the performance of the baseline model on each of the $5$ speakers. The average accuracy of the speakers at top-1 and top-5 is only $33.0\%$ and $51.6\%$ respectively. The accuracy of the current SOTA lipreading model on the LRW dataset is $88.5\%$ at top-1. This presents us with a huge scope for improvement. 

% Maybe add this line for direct inference.
% We do not show inference results directly on LRwP as the classes recorded by the patient are significantly different from the classes captured by LRW and for the classes common between those recorded by the patient and LRW, the accuracy is close to X at top 1 and Y at top 5 position. 
% Compared to the performance of the SOTA lipreading models on LRW reaching accuracy of 85\% at 1 on the dataset presents us with a huge scope for improvement. 

\textbf{Data Augmentation using Synthetic Data (Exp-\textit{cl$-$r$+$s})}: We augment the one-shot examples with a potentially unlimited number of synthetic examples. We train our models with varying amounts of synthetic data. 
% We plot a synthetic-real split graph as shown in Fig~\ref{fig:speaker1_synthetic} to determine the optimal amount of synthetic data needed for each speaker against the real examples. 
We plot a graph of accuracy against the combined synthetic and one-shot examples to determine the optimal amount of synthetic data needed for each speaker against the one-shot examples as shown in the supplementary, Fig.~1.
% the experiment shown in Fig~\ref{fig:speaker1_synthetic} for each speaker to find the optimal synthetic-real configuration. 
The optimal number of synthetic examples per speaker is reported in Table~\ref{tab:datasplit}. 

% We train two different models using LRwP and LTCN as backbones.
As shown in Table~\ref{tab:evaluation} Exp-\textit{cl$-$r$+$s}, we observe a significant jump in the accuracy consistently for every speaker with an overall improvement of $9\%$ and $7\%$ at top-$1$ and $5$, respectively on LRwP, and $7\%$ and $12\%$ at top-$1$ and $5$, respectively on LTCN. %With LTCN as the backbone, we observe similar results}. 
Although the accuracy improved over the baseline model, we observe that the model overfits on the synthetic data after a few epochs. Since we use only one example per class for the real domain, the model cannot foresee the target (real) variations that it may encounter during testing. We, therefore, try to introduce variations using Variational Encoders. 

\begin{figure*}
\begin{center}
\includegraphics[width=\linewidth]{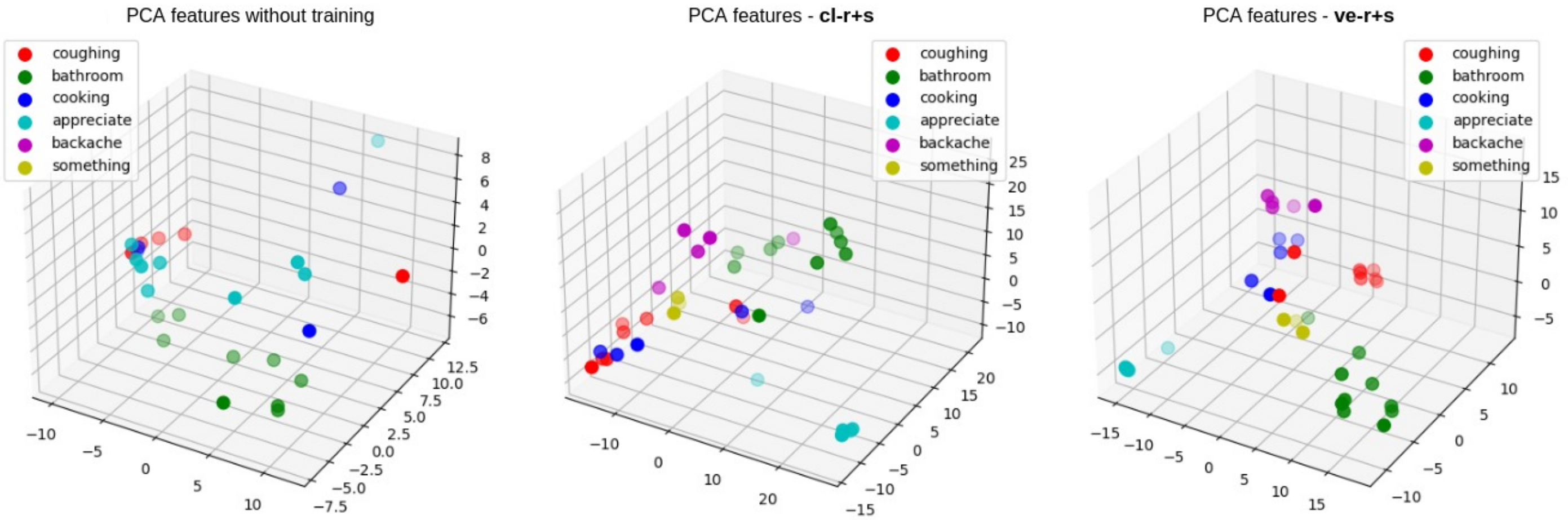}
\end{center}
   \caption{PCA visualization for the embeddings generated by the feature extraction layer of LRwP based encoder. (left) before training, (middle) trained on \textit{cl$-$r$+$s}, and (right) trained on\textit{ve$-$r$+$s}. Samples from the ALS patient's test set are used for visualization} % are collected from the test set of the ALS patient.}
%   \caption{LDA visualization of the embeddings generated by the Variational Encoder at the 0th, 70th, and 150th epoch in left to right order. Samples are collected from the test test of the ALS patient. Triangle and circle represents synthetic and real samples respectively. }
\label{fig:visualization}
\end{figure*}

\textbf{Variational Encoders (Exp-\textit{ve$-$r$+$s})}: 
%We use Variational Encoders to generate pseudo variations of the real examples. 
To train the network, we use the one-shot examples as the source domain. For the target domain, we combine the real and synthetic dataset. 
%This enables the model to fit on all the existing examples first while using the additional variations generated by the target encoder.
This lets the source encoder see the existing real examples and also become robust against the additional pseudo examples sampled from the target distribution.
We observed that using only the synthetic data in the source encoder makes the training highly unstable and the network does not converge. Instead, we allow the source encoder to first fit and then improve. 

As seen from Table~\ref{tab:evaluation} Exp-\textit{ve$-$r$+$s}, Variational Encoders consistently achieves the highest accuracy at top-5 across LRwP and LTCN. There is an overall improvement of $\sim19\%$ over the baseline and $\sim12\%$ over Exp-\textit{cl$-$r$+$s}.
%achieve the highest improvement in the top-5 accuracy across all the speakers, achieving an improvement of $17\%$ over the baseline and $12\%$ over Exp-2. 
The improvement at top-1 accuracy is comparatively much lower, $\sim15\%$ over the baseline and only $\sim6\%$ over Exp-\textit{cl$-$r$+$s}. We use PCA visualizations to analyze the latent representations learnt by the LRwP based source encoder, as shown in Fig.~\ref{fig:visualization} (right). We observe better separation for each class compared to the separation of Exp-\textit{cl$-$r$+$s} (middle). We also observe better disentanglement between non-homophenes such as `bathroom', `appreciate', and `coughing' compared to Exp-\textit{cl$-$r$+$s}. %where non-homophenes `bathroom', `backache', and `something' seem comparatively more entangled. 
For Varitional Encoders we observe that homophenes `coughing', `cooking', `something' are closer together. This suggests that the model trained on Variational Encoders can get confused for homophenes bringing the accuracy at top-1 down and at top-5 higher. For Exp-\textit{cl$-$r$+$s} however, all the classes seem equally apart. 
% \textcolor{red}{Need to edit this paragraph and also add LTCN scores and compare with other models}.

\textbf{Ad-hoc Data Augmentation (Exp-\textit{cl$-$r$+$s$+$aug} and Exp-\textit{ve$-$r$+$s$+$aug}):} In addition to the implicit variations introduced by Variational Encoders during the model training, we introduce explicit real-domain variations by augmenting the one-shot dataset. First, we use moviepy~\cite{moviepy} library to speed up and speed down the one-shot videos by a factor of 1.2$\times$ and 0.8$\times$.  The videos in our work are unconstrained, that is, the actual mouthing could be spread across several frames placed temporally anywhere in the video. Thus, we increase the video frames sequence length to 85 and add temporal variations during training by padding the videos with random number (between 0 to 20) of empty frames at the start and end. 
% We also increase the input sequence length to 85.} 
% increase the maximum sequence length of video frames from 64 to 85.}

We observe improvement in the performance across speakers for both, classification (Exp-\textit{cl$-$r$+$s$+$aug}) and Varitional Encoders (Exp-\textit{ve$-$r$+$s$+$aug}). We observe that for Spk-1 and Spk-2, the performance degrades. Upon further analysis, we find that the videos maintain constant pace and have fewer overall variations. Thus, the added variations during training behave as noise driving the performance down. The overall performance improves by 7\% and 3\% at top-1 on Exp-\textit{cl$-$r$+$s$+$aug} and Exp-\textit{ve$-$r$+$s$+$aug} over Exp-\textit{cl$-$r$+$s} and Exp-\textit{ve$-$r$+$s} respectively. We observe the best performance on one-shot setting with these ad-hoc additions for classification. For Variational Encoders, the improvements are less significant, especially at top-5, indicating the technique itself makes up for these ad-hoc additions.
%Although, the overall performance at top-5 drops minutely, the performance improvement for speakers Joe Biden, Spk-3, and the ALS patient is significant.} 
% Although, the model is now able to capture the class level characteristics from the synthetic data and speaker style of speaking from the real data, one-shot real examples is not enough to capture the underlying real domain variations of speaking a word. That is, a speaker will speak the exact same manner with different lip-movements, especially in the case of medical patients with limited facial muscle control. 
% We observe that homophenes such as `cooking' and `coughing' are aligned closer to each other. This suggests that the pseudo variations generated by the Variational Encoders could represent homophenes of the input one-shot examples. 

\textbf{Additional Experiments on the ALS Patient's Dataset}: To observe the affect of using different TTS models for generating synthetic data, we perform an ablation by eliminating one TTS model at a time for data generation and compare its performance against the data generated by combining both the TTS models. As shown in Table~\ref{tab:additional_evaluation}, the performance with both the TTS combined gives us the best performance. This indicates that the variations introduced by different TTS models are important for generalization. 

Lastly, we train an additional model with the additional data obtained from the patient's family to evaluate the performance difference between one-shot and two-shot setting. As seen from Table~\ref{tab:additional_evaluation}, the performance improves by $\sim7\%$ at top-5 against the best performing one-shot model on both training methodologies. Specifically for Variational Encoders, the performance improves to 71.64\% and 89.36\% at top-1 and top-5. This suggests that the network learns a better real-domain latent distribution using the extra real examples. 

\textbf{Evaluation on Joe Biden}: Although Joe Biden represents speakers without disabilities, he speaks in an American accent, while LRW is composed of British speakers. As can be observed from Table~\ref{tab:evaluation} Exp-\textit{cl$-$r}, the performance for Joe Biden on the baseline model is similar to the performance for the deaf speakers. A different accent can also be thought of as a different style of speaking. Thus the pretrained model does not directly adapt to Joe Biden. We observe in his case, adding synthetic data leads to significant improvements in the accuracy over the baseline with an average gain of $21\%$ at top-1 and top-5 (Exp-\textit{cl$-$r$+$s}). Variational Encoders fail to bring expected improvements compared to the other speakers. With LRwP, at top-1, the accuracy drops by $3\%$, and we see a marginal improvement at top-5.  We note that the TTS used to generate synthetic data is of an American accent. This suggests that the synthetic data captures the variations of the real domain exceptionally well in his case. Variational Encoders, on the other hand, adds more noise than valuable variations.

\section{Conclusion and Future Work}
In this work, we leap from previous lipreading approaches and propose a one-shot personalized lipreading framework to aid patients suffering from ALS or hearing disabilities. Due to the extreme scarcity of personalized data available for a medical patient, we generate synthetic data to augment our training process. We train our network with Variational Encoders, a domain adaptation technique, to bridge the gap between the synthetically generated examples and the available one-shot real examples for each class. Our method proves to be highly effective, and we achieve over $83\%$ top-5 accuracy for the ALS patient. %The patient currently regularly uses a web application to access the privately deployed model. 
We also report the performance of speaker-specific models trained for multiple speakers with hearing impairment and a speaker with no disability.
In the future, we would like to improve the accuracy of our model while also increasing the vocabulary it can handle. We believe our work achieves essential milestones for the lipreading community and can enhance the communication between people dependent on lipreading. It encourages research in the direction of few-shot lipreading that is an important real-world challenge that can have far-reaching applications including and beyond medical lipreading. Although, Variational Encoders has been proposed for few-shot lipreading, its utility can be explored in other areas with a few-shot settings. 

\textbf{Acknowledgement}: We would like to thank Anuraag Mullick and Vibha Mullick for their continual support in curating the dataset needed for this work. We also thank Harini Bhatt, who is the founder of ASRM Systems, for connecting us with several participants.

% We tackled the real-world medical challenge of lipreading disabled speakers in one-shot setting. We proposed a novel framework -- Variational Encoders -- 

\bibliography{bmvc}
\end{document}

\maketitle

%-------------------------------------------------------------------------

\section{Ablation Experiments}

In this section, we report ablation experiments to inspect and gain further insights on the different components of our network. Our experimental setup is identical to the setup mentioned in Section~3 of the main paper. For better readability during comparisons, we have copied the results of Table~2: Exp-\textit{cl$-$r} and Exp-\textit{ve$-$r$+$s} of the main paper to Table~\ref{tab:supp_evaluation}. 

We would like to also encourage the reader to view the supplementary video containing the results and comparisons.

\begingroup
\setlength{\tabcolsep}{7pt} % Default value: 6pt
\renewcommand{\arraystretch}{1} % Default value: 1
\begin{table}[H]
  \begin{center}
  \begin{tabular}{|| c || c | c | c | c || c | c | c | c || }
      \hline
      & \multicolumn{2}{ c }{cl$-$r} & \multicolumn{2}{| c ||}{ve$-$r} & \multicolumn{2}{ c | }{ve$-$r$+$s} & \multicolumn{2}{c || }{ve$-$r$+$s$-$mod}
      \\ \cline{2-9}
      Speaker & top1 & top5 & top1 & top5 & top1 & top5 & top1 & top5
      \\ \hline
      %\multirow{3}{*}{\makecell{Anuraag}} & Word-level &  3.6 & 4.3 & 4.7
      ALS patient & 49.3 & 62.6 & 51.2 & 65.4 & 66.4 & 81.4 & 64.6 & 74.9
      \\
      Spk-1 & 32.4& 52.1 & 36.9 & 48.9 & 48.6 & 68.5 & 48.3 & 60.3  %Benny
      \\ 
      Spk-2 & 23.1& 42.5 & 30.4 & 44.7 & 34.4 & 59.6 & 31.5 & 57.7 %Chen
      \\ 
      Spk-3 & 26.7 & 49.3 & 35.7 & 59.6 & 41.6  & 71.4 & 39.7 & 65.3 %Jazzy
      \\ 
      Joe Biden & 33.5 & 51.5 & 41.5 & 57.8 & 54.1 & 72.4  & 54.4 & 70.8
      \\ \hline
      Avg. & 33.0 & 51.6 & \textbf{39.1} & \textbf{55.2} & \textbf{49.0} & \textbf{70.6} & 47.7 & 65.8
      \\ \hline
  \end{tabular}
  \end{center}
  \caption{Evaluation of our models against each speaker in \%. All the metrics are evaluated on the curated test set made of only real-data. Exp-cl$-$r is the baseline trained on one-shot examples. Exp-ve$-$r denotes Variational Encoders trained on only one-shot examples. Exp-ve$-$r$+$s denotes Variational Encoders trained on synthetic and one-shot examples. Exp-ve$-$r$+$s$-$mod is Variational Encoders trained with no distance loss. All experiments are performed using LRwP as the backbone on real $+$ synthetic dataset. }
  \label{tab:supp_evaluation}
\end{table}
\endgroup

\begin{figure}[t]
\begin{center}
\includegraphics[width=0.8\linewidth]{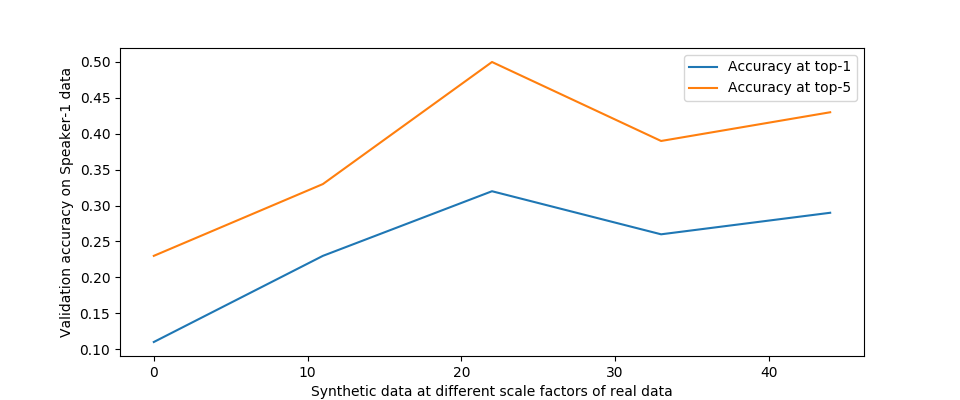}
\end{center} 
  \caption{Plot to determine the optimal amount of synthetic data against the real examples for Spk-1 at different scale factors of real data. Reference added in the main paper.}
\label{fig:graph}
\end{figure}

\subsection{Variational Encoders on Only Single Real Examples}

As reported in Section~3.1 of the main paper, our network is trained on the combined data of synthetic and one-shot examples. In this section, we remove the synthetic data to see its benefit. Instead of passing synthetic and one-shot examples, we only use one-shot examples as input to both encoders. Since the Variational Encoder itself acts as a data augmentation network, we assume the network to improve over the baseline (Table~\ref{tab:supp_evaluation} Exp-\textit{cl$-$r}). 
%Equation (4) of the main paper is used to train the network. 
% our network is trained using synthetic data and one-shot examples. In this section, we remove the synthetic data. We pass real data in both the source and the target encoders. Equation~(4) of the main paper is used to train the network. 

As shown in Table~\ref{tab:supp_evaluation} Exp-\textit{ve$-$r}, the performance of the reduced network improves by an average of $6\%$ and $4\%$ at top-1 and top-5 respectively over the baseline. However, the performance is still lower than the rest of the experiments mentioned in the main paper (Table~2 Exp-\textit{cl$-$r+s} and Exp-\textit{ve$-$r$+$s}). This suggests that Variational Encoders with just one example cannot surpass methods with additional data augmentation techniques. But, even with only one example, Variational Encoders is capable of bringing significant improvements to the model. 
% the reduced model is able to improve the performance of the baseline by an average of 6\% and 4\% respectively. Even though the performance drops compared to the other experiments mentioned in the main paper, we see that even with only one example, using our proposed approach can improve the performance of a model significantly compared to direct training only on the one-shot examples. 
Thus, our approach can be a potential data augmentation technique impacting other tasks requiring a low-data setting.
% the performance the reduced model is better than the reported baseline (Table~2, Exp-1 of the main paper) but it performs worse than all the other experiments. 

\subsection{Effect of $\Delta_{dist}$}

As reported in Equation~(4) of the main paper,
% given by, 
% \begin{equation}
%     \Delta_{ve} = \alpha \cdot (g(e_i, y) + g(\bar{z}_i, y)) + \gamma \cdot (| e_{x_i} - z_{\bar{x}_i} | - \beta \cdot D_{KL}( p(x_i) || p(\bar{x}_i))).
% \end{equation}
$\Delta_{ve}$ is a combination of $\Delta_{dist}$ that introduces domain invariance by reducing the distance between the feature space of source and target encoder, and $\Delta_{entropy}$ that allows the encoders to generate task-relevant latent embeddings and distributions using the loss of the downstream task. In this section, we remove $\Delta_{dist}$ and observe the behavior of the network using only $\Delta_{entropy}$. This allows us to determine the effect of introducing domain invariance. Loss of the reduced network is thus given by, 
\begin{equation}
    \Delta_{ve} = \Delta_{entropy} = \alpha \cdot (g(e_i, y) + g(\bar{z}_i, y))\ \{\alpha = 1\}
\end{equation}
% where $\alpha = 1$.

% We remove  $\Delta_{dist}$ to determine the effect of generating domain invariant features and reducing the distance loss. 
%Removing $\Delta_{dist}$ helps us estimate the contribution domain inavariance for the downstream task.
As can be seen from Table~\ref{tab:supp_evaluation} Exp-\textit{ve$-$r$+$s$-$mod}, the accuracy of the reduced network drops by $1\%$ and $5\%$ across the speakers at top-1 and top-5 respectively over Exp-\textit{ve$-$r$+$s}. 
%the accuracy of the trained model drops compared to our best performing model. There is an average accuracy drop of $1\%$ and $5\%$ across the speakers at top-1 and top-5 respectively. 
We observe, $\Delta_{entropy}$ plays a substantial role in our network. Generating variations using Variational Encoders lets the downstream classifier see various examples of the real domain making it robust. We observe the accuracy drop at top-5 is higher than top-1. The drop suggests that $\Delta_{dist}$ may introduce domain invariance across related mouth movements (such as homophenes). %This leads to related inferences at top-5 improving the accuracy at top-5.% which could also be occasionally detrimental to the downstream task. 

% The drop at top-1 accuracy is comparatively higher than top-5 which is in proportion with the improvements observed at top-1 and top-5 over the other models using our proposed approach. We can also infer that Variational Encoders generates variations across different features that can occasionally detrimental to the downstream task. 
% The major improvement over direct training (Exp 2, Table~ of the main paper) is observed from the entropy loss. This suggests that the entire architecture allows the downstream classifier to see more variations of the real examples which in turn helps the classifier become more robust.

\section{Need for Variational Encoders}

\subsection{Other Data Augmentation Techniques}

\begin{figure}
\begin{center}
\includegraphics[width=0.9\linewidth]{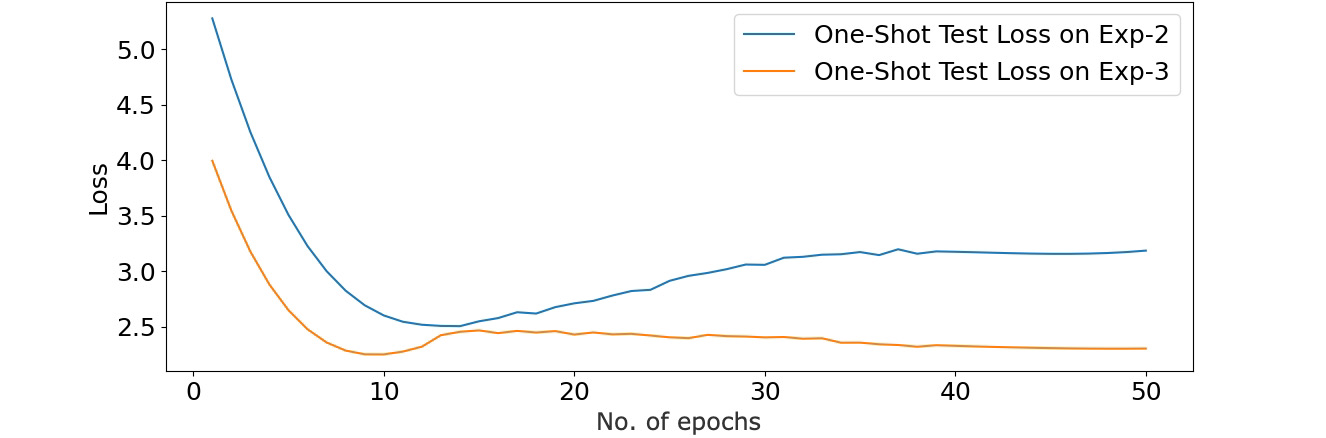}
\end{center} 
   \caption{Comparison between the test-loss of Exp-cl$-$r+s in the main paper with Exp-ve$-$r$+$s (our proposed approach) for the ALS patient.}
   %As can be seen from the graph, the test loss in Exp-cl$-$r+s saturates at Epoch 12, in Exp-ve$-$r$+$s, the loss slowly decreases.}
\label{fig:graph}
\end{figure}

In this section we report additional approaches that we tried to augment the one-shot examples. 
%Our proposed mechanism uses a lipsync model Wav2Lip to generate synthetic talking faces using the collected unlabeled videos and speech utterances generated using TTS models. 
An expert human lipreader can only lipread $40\%$ of a given sentence~\cite{cdc_2020}. Thus, manually labeling unlabeled videos is challenging. We therefore try to automatically label the unlabeled videos. 
%However, this is not a straightforward that as the unlabeled videos are recorded without any manual intervention. Thus they may not contain 
%This is necessary as the unlabeled videos are recorded without any manual intervention and may not even contain the words from the curated 200 classes.

\textbf{Best Match using Syncnet} - Syncnet~\cite{Chung16a} is an audio-video synchronization model to synchronize mouth movements and a given speech utterance. It generates a confidence score for a given audio and video segment. We exploit this method to find segments of unlabeled videos closely matching the generated TTS audios. We use different confidence thresholds to label the videos and vary the length of video segments based on the audio length. 

\textbf{Pseudo Labeling} - Pseudo Labeling~\cite{rpseudo, pseduolabel} is a technique that iteratively labels the unlabeled data in a semi-supervised setting. A model is trained on the labeled data and is used to label the unlabeled data known as pseudo labels. The model is trained further on the pseudo labels. These steps are repeated until the pseudo labels converge. It assumes the first model trained on the labeled data to be accurate so the pseudo labels are close to the actual labels. Pseudo labels can be noisy, but as the training progresses, labels converge to stable and accurate predictions. 
%We use this technique to label our data. 
We first segment the continuous unlabeled videos into smaller word-level video segments. We use lip-landmarks detected by dlib to measure the rate of change of lip-movements and determine the pauses between words. We then segment the videos on these pauses. We then iteratively label these videos using the pseudo labeling technique.

% is a technique used in a semi-supervised setting to label the unlabeled data, known as pseudo labels, using a model trained on simultaneously on labeled and pseudo labeled dataset. We used this approach to pseudo label our unlabeled videos. We first used a lip-landmark detection model to determine the rate of change of lip-positions and segmented the videos on estimated pauses. These segmented videos were then iteratively labeled using pseudo labeling. 

We observed that these methods led to noisy labels. The performance for each model mentioned Table~2 of the main paper dropped by an average of $15\%$ and $7\%$ on adding pseudo labels and syncnet labels respectively.
% Both of these methods led to noisy labels. We used the augmented data both in combination and without the synthetically generated data. We observed that in both the cases the performance of the existing models dropped. 

\subsection{Test Loss on One-Shot Examples}

Training a classifier on one-shot examples restricts the model from seeing many variations of the real domain and thus tends to underfit the real domain. We show this phenomenon in Fig.~\ref{fig:graph} that plots the test loss of Exp-\textit{cl$-$r+s} (of the main paper) and Exp-\textit{ve$-$r$+$s}. %Exp-\textit{cl$-$r+s} is a classification model trained on one-shot and synthetic examples. Exp-\textit{ve$-$r$+$s} is the proposed Variational Encoders model. 
In Exp-\textit{cl$-$r+s}, we see loss on the test set flattens and increases after Epoch-12. In Exp-\textit{ve$-$r$+$s} we see the loss keeps decreasing slowly. This suggests that Variational Encoders introduces meaningful variations of the real domain. %is successful in introducing meaningful variations of the real domain.

% As seen in Fig.~\ref{fig:graph}, the test loss on Exp-2, that directly trains a classifier on the combined dataset of synthetic and real examples, flattens and then increases after Epoch 12. Thus, the model does not fit well on the real test examples. This suggests that the one-shot examples do not capture the variations in the real domain well. The test loss on Exp-3 is much lower than Exp-3 and is slowly decreasing. This suggests that Variational Encoders can capture much more variations than the vanilla classifier. 

% \section{Deployed Website}

% \section{Failures}

\bibliography{egbib}